\documentclass[11pt,a4paper]{article}
\usepackage[hyperref]{acl2017}
\usepackage{times}
\usepackage{latexsym}
\usepackage{xspace}
\usepackage{amsmath, amsfonts, amssymb}
\usepackage{graphicx}
\usepackage{caption}
\usepackage{subcaption}
\usepackage{algorithm}
\usepackage{algpseudocode}
\usepackage{adjustbox}
\usepackage{multirow, arydshln, booktabs}
\usepackage{url}

\aclfinalcopy 

\newcommand{\method}[1]{\texttt{#1}\xspace}
\newcommand{\wordtovec}{\method{word2vec}}
\newcommand{\sg}{\method{skip-gram}}
\newcommand{\cbow}{\method{cbow}}
\newcommand{\doctovec}{\method{doc2vec}}
\newcommand{\dbow}{\method{dbow}}

\newcommand{\wdbow}{\method{w-dbow}}
\newcommand{\IDFdbow}{\method{w-dbow(IDF)}}
\newcommand{\CACNN}{\method{CA(CNN)}}
\newcommand{\CAGRU}{\method{CA(GRU)}}

\newcommand{\dataset}[1]{\textsc{#1}\xspace}
\newcommand{\wiki}{\dataset{wiki}}
\newcommand{\STS}{\dataset{sts}}

\newcommand{\var}[1]{\mathbf{#1}}
\newcommand{\set}[1]{\mathcal{#1}}
\newcommand{\step}[1]{\left<#1\right>}

\newcommand{\domain}[1]{{\textsl{#1}}\xspace}
\newcommand{\best}[1]{\textbf{#1}}

\title{Context Aware Document Embedding}

\author{Zhaocheng Zhu \\
  School of Electronics Engineering \\
  and Computer Science, \\
  Peking University \\
  {\tt zhaochengzhu@pku.edu.cn} \\ \And
  Junfeng Hu \\
  Key Laboratory of \\
  Computational Linguistics, \\
  Ministry of Education, \\
  Peking University \\
  {\tt hujf@pku.edu.cn} \\}

\date{}

\begin{document}
\maketitle
\begin{abstract}
Recently, \doctovec has achieved excellent results in different tasks \cite{Lau;2016}.
In this paper, we present a context aware variant of \doctovec. We introduce a
novel weight estimating mechanism that generates weights for each word
occurrence according to its contribution in the context, using deep neural
networks. Our context aware model can achieve similar results compared to \doctovec
initialized by Wikipedia trained vectors, while being much more efficient and
free from heavy external corpus. Analysis of context aware weights shows they
are a kind of enhanced IDF weights that capture sub-topic level keywords in
documents. They might result from deep neural networks that learn hidden
representations with the least entropy.

\end{abstract}

\section{Introduction}

Knowledge representation, as a critical prerequisite for many machine learning
tasks, has always been a central problem in the field of natural language
processing (NLP). As for the representation of documents, an established form
is to use bag-of-words (BOW) or term frequency-inverse document frequency
(TF-IDF) representations. Another widely adopted method is generative topic
models, such as latent semantic analysis (LSA) \cite{Deerwester;1990} and
latent dirichlet allocation (LDA) \cite{Blei2003}.

Recently, \newcite{Bengio;2003} proposed a window-based unsupervised word embedding
method. Following his approach, \newcite{Mikolov;2013a} introduced two new
log-linear models, \sg and \cbow. \newcite{Mikolov;2013b} gave a highly efficient
implementation of those two models, and distributed it as \wordtovec, which has
been widely used as a tool in language related tasks.

Inspired by the success of \wordtovec, \newcite{Le;2014} extended \wordtovec into
\doctovec, which produces a vector representation for each document, known as
"document embedding". \newcite{Dai;2015} further examined \doctovec and found
analogy features on Wikipedia (e.g. "Lady Gaga" - "American" + "Japanese" $\approx$
"Ayumi Hamasaki"). However, others have struggled to reproduce such results.
Most recently, \newcite{Lau;2016} made an empirical evaluation of \doctovec,
and revealed its potential on different tasks.

Although \doctovec has produced promising results, we doubt its basis as it
implicitly assigns the same weight to each word occurrence when training document
vectors. This is counter-intuitive, since human never give equal attention to
different parts of a sentence. Consider the following sentence as an example:

\begin{quote}
\small
\begin{verbatim}
There are many activities
including but not limited to
running, jumping, and swimming.
\end{verbatim}
\end{quote}

When reading this sentence, we are not concerned about the "there be" term. We
will probably ignore "limited to", because "including" and "but not" indicate
the parenthesis character of that term. The sentence can still be understood
even if some parts are missing:

\begin{quote}
\small
\begin{verbatim}
... many activities including
... running, jumping, ...
\end{verbatim}
\end{quote}

Motivated by such facts, we propose a context aware document embedding based on
\doctovec. Our method takes a novel approach that estimates weights for each
word occurrence by measuring the shift of the corresponding document vector if
the word is substituted by another. We use convolutional neural networks (CNN)
and gated recurrent units (GRU) as auxiliary models for the space of document
vectors. We compared our model with benchmarks in \cite{Lau;2016} and \doctovec
with IDF weights to show the advantage of our model. To give a convincing
illustration of our model, we visualized the hidden states of deep neural
networks. Our findings suggest that context aware weights are a kind of
enhanced IDF weights that are especially good at capturing sub-topic level
keywords in documents, since neural networks can substantially extract
asymmetric context features, despite trained with unsupervisedly embedded
targets.

\section{Related Work}

\subsection{Distributed Bag of Words}

The departure point of the context aware model is the distributed bag of word
(DBOW) model of \doctovec proposed in \cite{Le;2014} trained with the negative
sampling procedure \cite{Mikolov;2013b}. \footnote{This paper uses the gensim
implementation of \doctovec \cite{Rehurek;2010}} \dbow uses a similar fashion
like \sg \cite{Mikolov;2013a} to train a document vector ($\var d^i$) for each
document with its context word vectors ($\var c^i_j$). By adopting negative
sampling procedure, the objective is to maximise the likelihood of
$P(\var c^i_j|\var d^i)$ while minimising the likelihood of $P(\var c'|\var d^i)$,
where $\var c'$ is a random sample \cite{Goldberg;2014,Lau;2016}. Therefore,
the objective function is given as:

\begin{equation}
    \sum_j{\left[\log\sigma({\var c^i_j}^\intercal \var d^i) - \\
    \sum_{k=1, \var c' \sim P_n(c)}^n{\log\sigma({\var c'}^\intercal \var d^i)}\right]}
    \label{eq:dbow-loss}
\end{equation}

where $\sigma$ is the sigmoid function, $n$ is the number of negative samples,
and $P_n(c)$ is a distribution derived from term frequency.

Despite that \dbow works with randomly initialized context word vectors, it is
suggested that the quality of embeddings is improved when context word vectors
are jointly trained by \sg and \dbow \cite{Dai;2015,Lau;2016}. It is also
observed that by initializing word vectors from pre-trained \wordtovec of large
external corpus like \wiki, the model converges faster as well as performs
better \cite{Lau;2016}.

\subsection{CNN}

\begin{figure}
    \includegraphics[width=.5\textwidth]{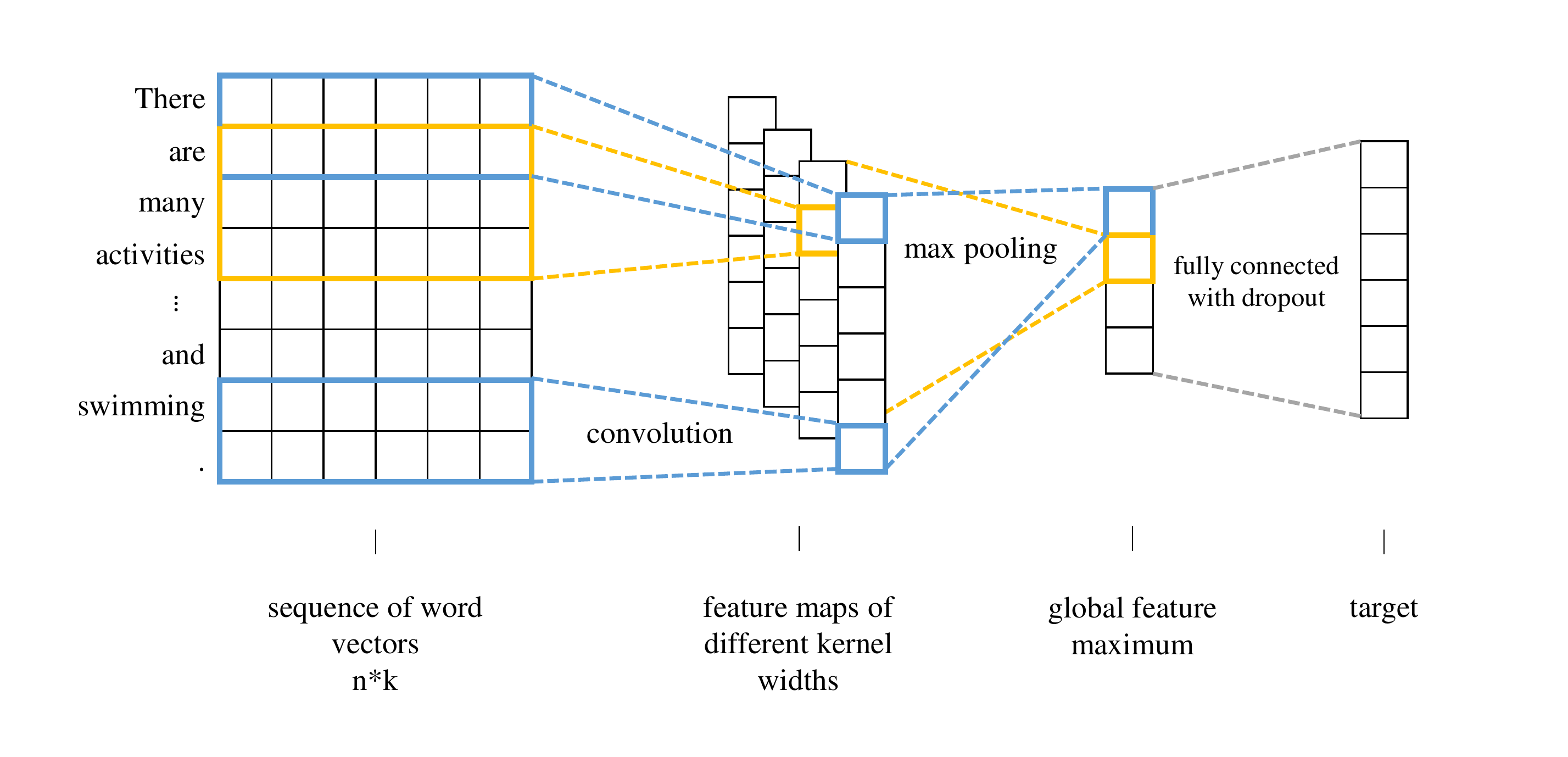}
    \caption{CNN architecture}
    \label{fig:CNN-arch}
\end{figure}

Convolutional neural networks (CNN) proposed by \newcite{LeCun;1998} are a kind
of partially connected network architecture. CNN exploits convolution kernels to
extract local features and uses pooling method to agglomerate features for
subsequent layers, which have achieved excellent results in many NLP tasks
\cite{Collobert;2011,Yih;2014}. A typical CNN architecture for NLP consists of
a convolutional layer, a global max pooling layer, and a fully connected layer
, as is shown in Figure \ref{fig:CNN-arch} \cite{Kim;2014}. \footnote{All neural
networks are implemented in Keras with Tensorflow backend\cite{Chollet;2015}
\label{fn:keras}}

For a document, the model takes a sequence of word vectors as input, and
applies convolution along the sequence. The kernels are designed with different
widths to extract features of different scales. Following each feature map,
there is a max-pooling, which leads to a global maximum of the feature over the
sequence. Finally, all features are concatenated and fully connected to the
target, for either regression or classification purpose.

\subsection{GRU}

\begin{figure}
    \includegraphics[width=.5\textwidth]{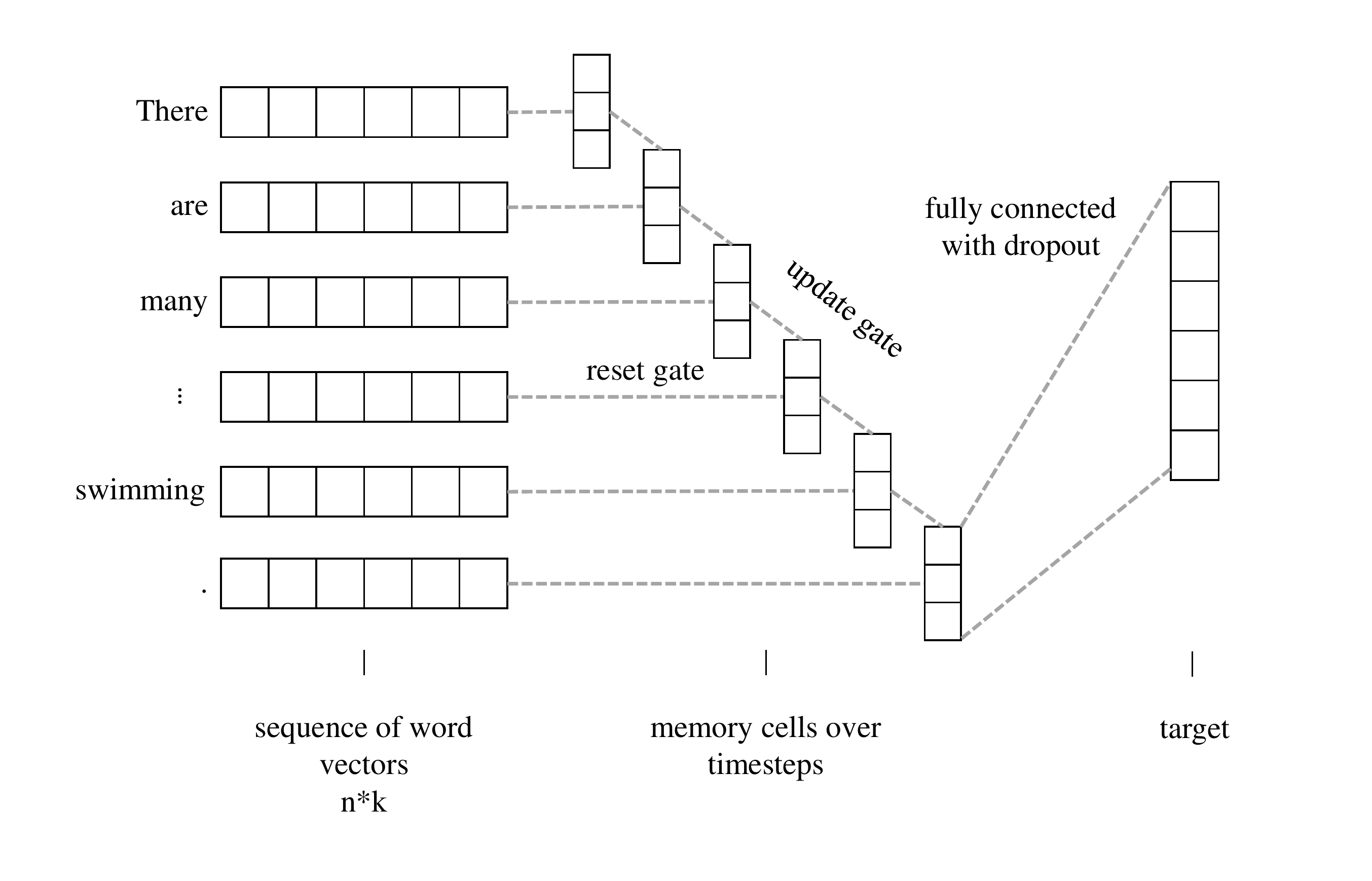}
    \caption{GRU architecture}
    \label{fig:GRU-arch}
\end{figure}

Gated recurrent units (GRU), introduced by \newcite{Cho;2014a,Cho;2014b,Chung;2014},
is a kind of recurrent neural networks (RNN) often used in semantics related
tasks. An RNN is a neural network that scans input items one by one, and produces
a representation for every prefix of the sequence, which can be formally written
as

\begin{equation*}
    \var h_{\step t}=f(\var h_{\step{t-1}}, \var x_t)
\end{equation*}

where $\var x_t$ is the $t^{th}$ item, $\var h_{\step t}$ is the hidden
representation at timestep $t$, and $f$ is a non-linear activation function. In
GRU, a reset gate $\var r_{\step t}$ and an update gate $\var z_{\step t}$ are
used for regulating $f$.

\begin{equation*}
    \var z_{\step t}=\sigma(\var W_z\var x_t+\var U_z\var h_{\step{t-1}})
\end{equation*}

\begin{equation*}
    \var r_{\step t}=\sigma(\var W_r\var x_t+\var U_r\var h_{\step{t-1}})
\end{equation*}

where $\sigma$ is a non-negative activation function, usually sigmoid or
hard sigmoid function. Then $f$ is computed by

\begin{equation*}
    f(\var h_{\step{t-1}}, \var x_t)=\var z_{\step t}\odot \var h_{\step{t-1}}+(1-\var z_{\step t})\odot\tilde{\var h}_{\step t}
\end{equation*}

with

\begin{equation*}
    \tilde{\var h}_{\step t}=\phi\left(\var W\var x_t+\var U(\var r_{\step t}\odot \var h_{\step{t-1}})\right)
\end{equation*}

where $\phi$ stands for $tanh$ function.

The two-gate design enables GRU to represent both long-term and short-term
dependencies over timesteps. Units with update gate frequently activated
tend to capture long-term dependencies, while units with reset gate activated
tend to capture short-term dependencies \cite{Cho;2014a}. For regression targets
over the sequence, a fully connected layer with dropout is added to the last
timestep, as Figure \ref{fig:GRU-arch} shows. \textsuperscript{\ref{fn:keras}}

\section{Context Aware Model}

Our context aware model is an unsupervised model that trains \doctovec with
word occurrence weights regarding their contexts. To generalize the problem,
we will first define weighted \dbow (notated as \wdbow), then illustrate
the context aware model as a specific one of \wdbow.

A \wdbow model is defined as a triplet $(\set D, \set C, \set W)$. In the
triplet, $\set D$ and $\set C$ are documents and context words for the vanilla
\doctovec model, respectively. $\set W$ is a set of weights corresponding to
each word occurrence $\var c\in \set C$. Similar to Equation (\ref{eq:dbow-loss}),
the target is to minimize the following function:

\begin{equation}
    \sum_j{\var w^i_j\left[\log\sigma({\var c^i_j}^\intercal \var d^i) -
    \sum_{k=1, \var c' \sim P_n(c)}^n{\log\sigma({\var c'}^\intercal \var d^i)}\right]}
    \label{eq:wdbow-loss}
\end{equation}

where $\var w^i_j\in \psi(\set W)$. Here we introduce a normalizing function
$\psi$ on $\set W$, because $\set W$ may be scaled or biased in non-trivial cases.
We adopt a global temperature softmax for $\psi$, which means

\begin{equation*}
    \psi(\var w^i_j)=\frac{|\set W|e^{\var w^i_j/T}}{\sum_{\var w\in \set W}{e^{\var w/T}}}
\end{equation*}

$T$, known as the "temperature", is a hyperparameter that controls the softness
of a softmax function. The result is scaled by $|\set W|$ times so that the
average $\var w$ is $1$, which is the case in \dbow. Therefore, hyperparameters
like learning rate from \dbow can be applied to \wdbow directly.

\begin{table*}
    \center
    \begin{adjustbox}{max width=0.95\textwidth}
        \begin{tabular}{cccccc}
            \toprule
            \textbf{Input Size} & \textbf{\#Timesteps} & \textbf{Output Size} & \textbf{Loss}   & \textbf{Optimizer} &
            \textbf{Epoch} \\
            \midrule
            300                 & 93                  & 300                  & cosine distance  & adam(lr=0.001)     &
            100 \\
            \bottomrule
        \end{tabular}
    \end{adjustbox}
    \caption{Shared hyperparameters of auxiliary models for STS task}
    \label{tab:shared-params}
\end{table*}

The context aware model is a \wdbow that generates weights in the following way.
First, a vanilla \dbow is trained on the corpus, and then weight $\var w^i_j$
is computed by randomly substitute $\var c^i_j$ and measuring the shift in
document vector $\var d^i$ with cosine distance. This makes sense because there
will be little shift if the word can be inferred from its context, otherwise a
large shift should take place, as replacing the word impedes the meaning of the
document. To predict document vectors on new word sequences, we trained an
auxiliary model to regress document vectors with word vector sequences. We
utilize CNN (Figure \ref{fig:CNN-arch}) and GRU (Figure \ref{fig:GRU-arch}) for
this task. Hence, the corresponding context aware models are named as \CACNN
and \CAGRU. Algorithm \ref{alg:weights} gives details of this procedure

\begin{algorithm}[h]
    \caption{Generate context aware weights}
    \begin{algorithmic}[1]
        \State $aux\_model \gets$ CNN or GRU
        \State $aux\_model.train(word\_sequence, doc)$
        \For{$i \gets 1$ to $|docs|$}
            \For{$j \gets 1$ to $|word\_sequence^i|$}
                \State $shifts \gets \varnothing$
                \For {$k \gets 1$ to $sample\_count$}
                    \State $s' \gets word\_sequence^i$
                    \State $s'_j \gets random\_word()$
                    \State $d' \gets aux\_model.predict(s')$
                    \State $d \gets doc^i$
                    \State $shifts.add(cos\_distance(d', d))$
                \EndFor
                \State $\var w^i_j \gets average(shifts)$
            \EndFor
        \EndFor
    \end{algorithmic}
    \label{alg:weights}
\end{algorithm}

The weight is computed by averaging all the shifts of random substitutions. The
random word is sampled from the global vocabulary with probability proportional
to term frequency (TF). We also propose an economic way that take samples from
words sharing the same part-of-speech (POS). Experiments show that two methods
can achieve similar results, as will be shown in Section \ref{sec:experiments}.

\section{Experiments}
\label{sec:experiments}

\subsection{Evaluation methods}
We conduct experiments on the semantic textual similarity (STS) task. It is a
shared task held by SemEval \cite{Agirre;2015}. In STS, the goal is to
automatically predict a score for each sentence pair according to their semantic
similarity. The result is evaluated by computing the Pearson correlation
coefficient between the predicted score and the ground truth.

Since the official dataset provided by SemEval is quite small, we combine all
datasets from 2012 to 2015 as the training set, following the approach in
\cite{Lau;2016}. We take the \domain{headlines} domain of 2014 as the
development set, and test on the 2015 dataset. It is reliable to specify the
test set as a subset of the training set, as both our methods and baselines are
unsupervised.

All datasets are preprocessed by lowercasing and tokenizing the sentence, using
Stanford CoreNLP \cite{Manning;2014}. POS tags are also obtained through
Stanford CoreNLP.

\subsection{Baselines}

To demonstrate the advantages of context aware models, we compare them with
unsupervised baselines proposed in \cite{Lau;2016}, including a linear
combination of word vectors from \sg, and several \dbow with different training
settings. The linear combination computes a document embedding for a sentence
by averaging over all its context with word vectors trained on the full
collection of English Wikipedia entries. \footnote{Using the pretrained model
by \cite{Lau;2016}: \url{https://github.com/jhlau/doc2vec} \label{fn:pretrained}}

For \dbow, we train one model on the STS dataset. \footnote{We use the same
hyperparameters for all \dbow and \wdbow trained on the STS dataset: vector
size = 300, window size = 15, min count = 1, sub-sampling threshold = $10^{-5}$,
negative sampling = 5, epoch = 400. \label{fn:dbow-hyperparams}} In the
following context, we will refer these datasets as \wiki and \STS respectively.
We train another model on \wiki and exploit it to infer document vectors for
\STS without updating any hidden weights. \textsuperscript{\ref{fn:pretrained}}

As it is observed in practice that using pretrained word vectors from external
corpus can benefit the performance of \dbow, we experiment a third \dbow
baseline with word vectors initialized by \sg trained on \wiki.
\textsuperscript{\ref{fn:pretrained},}\footnote{We test on both trainable and
untrainable word vector initialization for \dbow, and receive negligible
difference. Therefore, we only list the result of the trainable version.}

Our context aware models are concrete implementations of \wdbow. Because \wdbow
is consistent with \dbow, we force context aware models to use the same
hyperparameters as \dbow. \textsuperscript{\ref{fn:dbow-hyperparams}} We only
optimise hyperparameters of the auxiliary model (i.e. CNN or GRU) towards the
target of document vectors through cross-validation on the training set, as
well as the temperature $T$ on the development set. In the following context,
these two methods are referred as \CACNN and \CAGRU respectively.

To distinguish context aware models from other weighting methods, we also
introduced a \wdbow baseline with weights from inverse document frequency (IDF)
as \IDFdbow. We optimise its temperature $T$ separately from context aware
models.

\subsection{Experiments}

\begin{table*}
    \center
    \begin{adjustbox}{max width=0.95\textwidth}
        \begin{tabular}{ccc@{\hspace{20pt}}ccccc}
            \toprule
            \multirow{2}{*}{\textbf{Domain}} & \textbf{\sg} & \multicolumn{3}{c}{\textbf{\dbow}} & \textbf{\CACNN}
            & \textbf{\CAGRU} & \textbf{\IDFdbow} \\
                                             & \wiki        & \STS  & \wiki & \STS(\wiki init)   & \multicolumn{2}{c}{\STS}
                              & \STS              \\
            \midrule
            \domain{answers-forums}           & 0.516        & 0.647 & 0.666 & \best{0.675}       & 0.670
            & 0.662           & 0.656 \\
            \domain{headlines}               & 0.731        & 0.768 & 0.746 & 0.782              & 0.785
            & 0.787           & \best{0.788} \\
            \domain{answers-students}         & 0.661        & 0.640 & 0.628 & 0.654              & \best{0.683}
            & 0.676           & 0.660 \\
            \domain{belief}                  & 0.607        & 0.764 & 0.713 & \best{0.773}       & 0.772
            & 0.764           & 0.760 \\
            \domain{images}                  & 0.678        & 0.781 & 0.789 & \best{0.800}       & 0.793
            & 0.793           & 0.787 \\
            \bottomrule
        \end{tabular}
    \end{adjustbox}
    \caption{Results over STS task with different unsupervised methods}
    \label{tab:method-compare}
\end{table*}

\begin{table*}
    \begin{center}
        \begin{adjustbox}{max width=0.95\textwidth}
            \begin{tabular}{cc@{\hspace{15pt}}c@{\hspace{20pt}}c@{\hspace{15pt}}c}
                \toprule
                \multirow{2}{*}{\textbf{Domain}} & \multicolumn{2}{c}{\textbf{\CACNN}}
                                                 & \multicolumn{2}{c}{\textbf{\CAGRU}} \\
                                                 & global   & POS      
                                                 & global   & POS      \\ 
                \midrule
                \domain{answers-forums}           & 0.663     & \best{0.670} 
                                                 & 0.668  & 0.662        \\ 
                \domain{headlines}               & 0.786     & 0.785        
                                                 & 0.786  & \best{0.787} \\ 
                \domain{answers-students}         & 0.673     & \best{0.683} 
                                                 & 0.675  & 0.676        \\ 
                \domain{belief}                  & 0.760     & \best{0.772} 
                                                 & 0.761  & 0.764        \\ 
                \domain{images}                  & \best{0.800} & 0.793     
                                                 & 0.792  & 0.793        \\ 
                \bottomrule
            \end{tabular}
        \end{adjustbox}
        \caption{Context aware models with random sampling from different distributions.
        The global distribution is proportional to TF of each word. The POS distribution
        is proportional to TF of each word grouped by POS. Samples are taken from the
        corresponding distribution regarding POS of the substituted word. We take 50
        and 10 samples for global distribution and POS distribution respectively.}
        \label{tab:CA-compare}
    \end{center}
\end{table*}

First, we decide hyperparameters for our auxiliary models. For CNN, it is
observed through cross-validation that kernels with width from 3 to 8 best
capture features from the word vector sequence to make the vanilla document
vector. Considering the scale of \STS dataset, we use 128 kernels for each
width. For GRU, we use 512 hidden units. Both deep networks have 1.4M trainable
parameters approximately, indicating they should have the same capability.


Then we optimise the temperature $T$ for \wdbow models. We find that results on
the development set almost follows a unimodal function as $T$ varies, which
facilitates the optimization of $T$ a lot. In general, $T$ around 1/15--1/14
works for both \CACNN and \CAGRU, while $T$ around 5--6 works for \IDFdbow.

Experiments show that our context aware models outperforms \dbow in all 5
domains (Table \ref{tab:method-compare}). Of two purposed models, \CACNN works
a little better and results at the same level of \dbow initialized by word
vectors from \wiki. Besides, \IDFdbow also gets a better performance than \dbow,
which buttresses the consistency of our definition for \wdbow. More
interestingly, all \wdbow baselines make excellent results in the domain of
\domain{answers-students}, compared to any \dbow approach. We will give a
detailed analysis of this in Section {\ref{sec:introspection}}.

We also compared \CACNN and \CAGRU with different word distributions. Table
\ref{tab:CA-compare} gives results in detail. Consistent with our estimation,
the difference between global distribution and POS distribution is not
significant. Therefore, gain of context aware models comes from weighting
method rather than sampling from POS distribution. However, we recommend to use
POS distribution for random substitution, as it is much smaller and more
efficient.

\section{Model Introspection}
\label{sec:introspection}

\begin{figure*}
    \begin{center}
        \begin{adjustbox}{max width=\textwidth}
            \begin{subfigure}{0.2\textwidth}
                \includegraphics[width=0.99\textwidth]{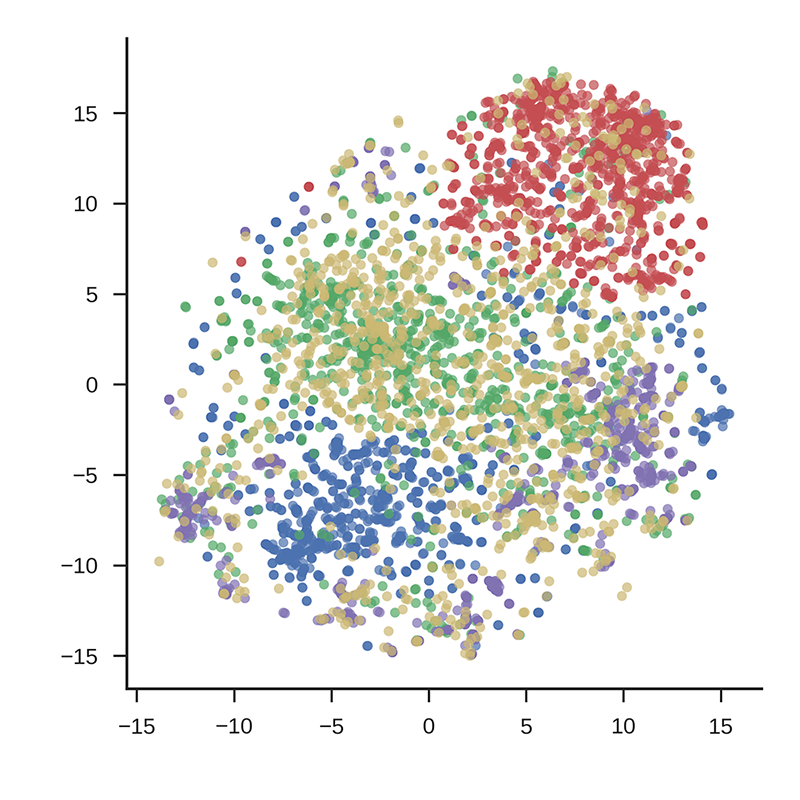}
                \caption{\dbow}
            \end{subfigure}
            \begin{subfigure}{0.2\textwidth}
                \includegraphics[width=0.99\textwidth]{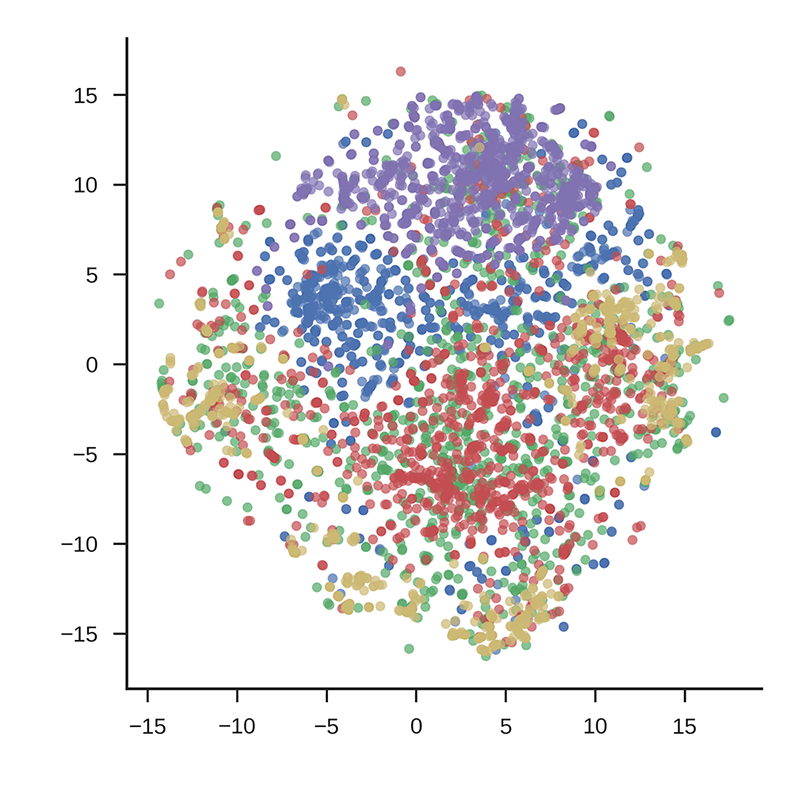}
                \caption{\dbow(\wiki init)}
            \end{subfigure}
            \begin{subfigure}{0.2\textwidth}
                \includegraphics[width=0.99\textwidth]{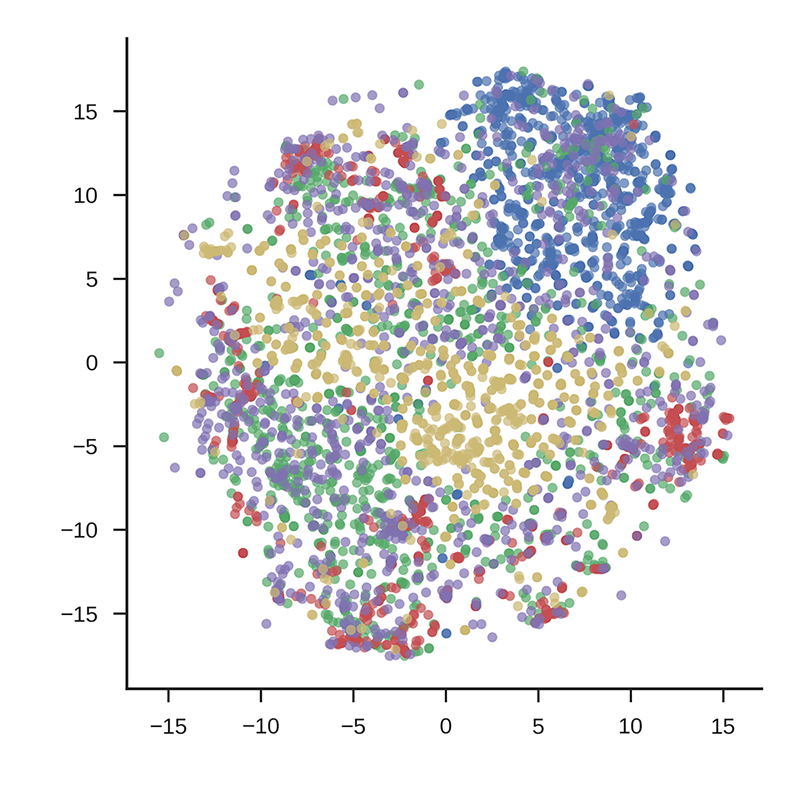}
                \caption{\CACNN}
            \end{subfigure}
            \begin{subfigure}{0.2\textwidth}
                \includegraphics[width=0.99\textwidth]{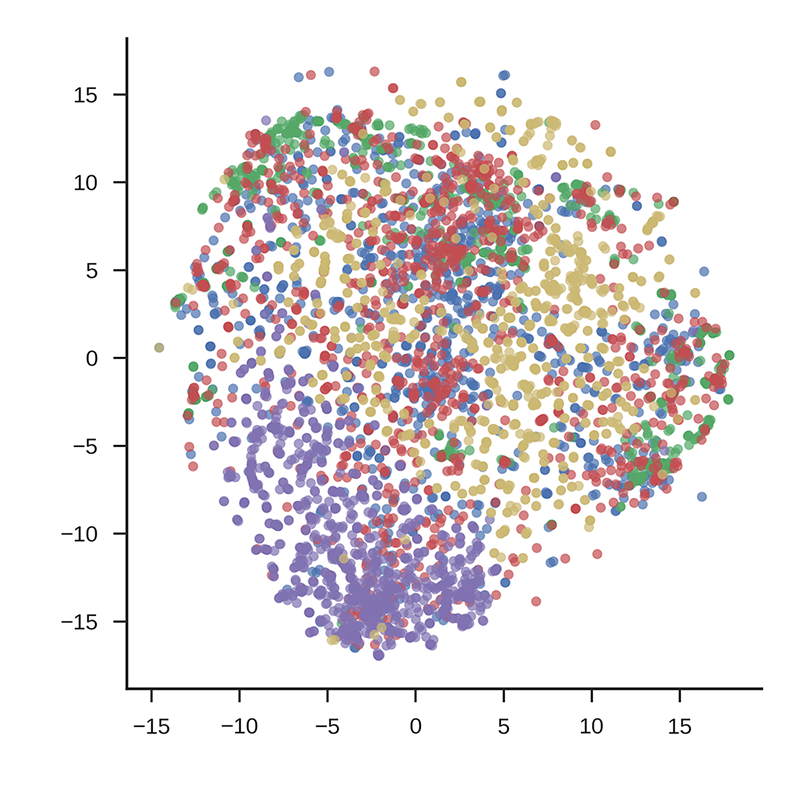}
                \caption{\CAGRU}
            \end{subfigure}
            \begin{subfigure}{0.2\textwidth}
                \includegraphics[width=0.99\textwidth]{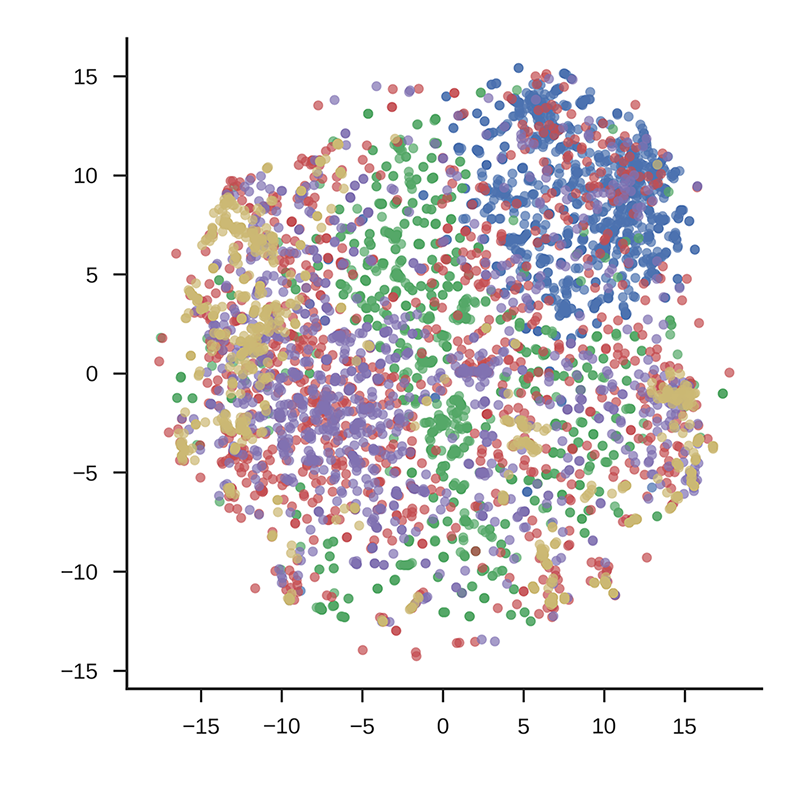}
                \caption{\IDFdbow}
            \end{subfigure}
            \begin{subfigure}{0.15\textwidth}
                \includegraphics[width=0.99\textwidth]{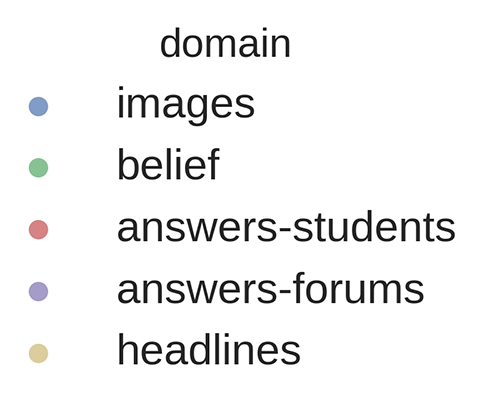}
            \end{subfigure}
        \end{adjustbox}
        \caption{Distribution of different document vectors under t-SNE}
        \label{fig:vec-distributions}
    \end{center}
\end{figure*}

\begin{figure*}
    \includegraphics[width=.95\textwidth]{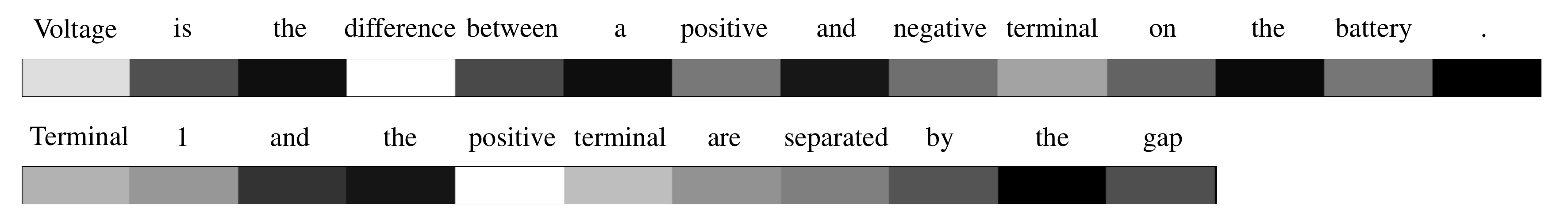}
    \caption{A sample of weights learned by the context aware model. Brighter bars denote larger weights.}
    \label{fig:weight-sample}
\end{figure*}

Being unsupervised models, our context aware models surpass the vanilla \dbow,
and even show some advantage towards \dbow(\wiki init). This is momentous
because context aware models only exploit the local corpus, without external
priori like pretrained vectors. Hence, we believe context aware models do
extract more information from the corpus. We will shed some light on why and
how context aware models work by looking into document vectors and hidden
representations of the model.

In context aware models, weights learned by word-level substitution consists of
two components. One is a base value for each word, the other is a
context-related bias for each word occurrence. The base value is something close
to IDF, as there is a moderate correlation between context aware weights and
IDF. Hence, context aware models can be viewed as an enhanced version of
\IDFdbow. \footnote{Typically, the Pearson's r between context aware weights
and IDF is 0.5--0.6.}

Generally, both context aware models and \IDFdbow give equal or smaller
similarity answers in STS task. They tend to distinguish sentences rather than
find trivial similarities. In other words, IDF weighted loss makes document
embeddings insensitive to common words. Context aware models even enable them
to neglect common words given the context in a self-adaptive manner, which is
very similar to lateral inhibition in cognition and neuroscience.

To illustrate this, consider the distribution of document vectors on the test
set using t-SNE \cite{Maaten;2008} in Figure \ref{fig:vec-distributions}.
Document vectors of different domains are marked with different colors. Since
domains are natural categories, dots of the same color should gather as a
cluster. Consistent with our experience, \dbow forms good clusters. However,
none of three \wdbow models gives such a clear result. In context aware
embeddings, though clusters still can be observed, a number of dots are
scattered far away from their centers. The phenomenon is extremely significant
in \domain{answers-students} (red), where \dbow performs worst and context
aware models improve most. This might be attributed to that different from
coarse-grained task like clustering, STS task more relies on fine-grained
features, where our models have advantage over vanilla \dbow.

We then examine the domain of \domain{answers-students}. It contains students
answer to electricity problems, of which most share the same topic, but their
semantic similarity varies. As for a concrete example, we randomly pick a
sentence pair from \domain{answers-students} whose similarity differs much in
\dbow and context aware models. Figure \ref{fig:weight-sample} clearly shows
weights learned by \CACNN. It can be spotted that low weights are assigned to
common words. Moreover, context aware model neglects jargons like "voltage" or
"terminal", but focuses on "difference", which is a relatively rare occurrence
given the context. In fact, it is word like "difference" and "positive" that
defines the key point of a sentence in a topic. With context aware weights,
their contribution are well amplified.

\begin{table}[htb]
    \begin{center}
        \begin{adjustbox}{max width=0.45\textwidth}
            \begin{tabular}{ccc@{\hspace{20pt}}c}
                \toprule
                \multirow{2}{*}{\textbf{Context word}} & \textbf{CNN} & \textbf{GRU} & \textbf{IDF} \\
                                    & \small{\#feature change} & \small{norm of $\var r_{\step t}$} \\
                \midrule
                Voltage                                & 393          & 5.037        & 5.805       \\
                is                                     & 157          & 8.847        & 1.526       \\
                the                                    & 98           & 10.168       & 0.877       \\
                difference                             & 385          & 7.153        & 5.925       \\
                between                                & 200          & 8.957        & 4.678       \\
                a                                      & 96           & 10.039       & 0.953       \\
                positive                               & 324          & 5.916        & 4.383       \\
                and                                    & 99           & 9.334        & 1.671       \\
                negative                               & 322          & 6.017        & 5.363       \\
                terminal                               & 328          & 5.393        & 4.066       \\
                on                                     & 122          & 10.349       & 2.171       \\
                the                                    & 53           & 10.476       & 0.877       \\
                battery                                & 230          & 6.889        & 3.422       \\
                .                                      & 15           & 10.570       & 0.417       \\
                \midrule
                \textbf{Correlation to IDF}            & 0.915        & -0.845       & 1           \\
                \bottomrule
            \end{tabular}
        \end{adjustbox}
        \caption{Influence of each context word on hidden representations}
        \label{tab:hidden-states}
    \end{center}
\end{table}

Notice that auxiliary models are trained with document vectors, which are
generated according to context words homogeneously. But shifts in the vector
space do behave heterogeneously regard to each substituted occurrence. To find
the origin of such asymmetry, we investigate on the hidden states of auxiliary
models. For CNN, we count the number of global features that change in
substitution. For GRU, since the hidden representation varies in different
timesteps, we count the norm of $\var r_{\step t}$, as it implies how much
units "decline" the input. Surprisingly, both auxiliary models reveal unequal
hidden states, as shown in Table \ref{tab:hidden-states}. The number of
features that a word contributes to in CNN is highly correlated with its IDF,
while the extent to which hidden units turn down a word in GRU is highly
negatively correlated with its IDF. Since IDF is the entropy of every word
given its distribution over a corpus, we are inclined to believe that the first
layer of deep neural networks learns hidden representations towards the least
entropy over the corpus (i.e. maximize the use of hidden units).

\section{Conclusion}

We introduce two context aware models for document embedding with a novel
weight estimating mechanism. Compared to vanilla \dbow method, our approach
infers a weight for each word occurrence with regard to its context, which
helps document embedding to capture sub-topic level keywords. This property
facilitates learning a more fine-grained embedding for semantic textual
similarity task as well as eases training on large external corpus. We claim
that context aware weights is composed of an IDF base and a context-related
bias. This might be induced by deep neural networks as they naturally learns
representations with the least entropy, even when the target is generated
homogeneously.

\nocite{Pennington;2014}
\nocite{Zhang;2017}

\bibliography{paper}
\bibliographystyle{acl_natbib}
\appendix

\end{document}